# Polysemy in Controlled Natural Language Texts

Normunds Gruzitis and Guntis Barzdins

Institute of Mathematics and Computer Science, University of Latvia
normundsg@ailab.lv, guntis@latnet.lv

**Abstract**. Computational semantics and logic-based controlled natural languages (CNL) do not address systematically the word sense disambiguation problem of content words, i.e., they tend to interpret only some functional words that are crucial for construction of discourse representation structures. We show that micro-ontologies and multi-word units allow integration of the rich and polysemous multi-domain background knowledge into CNL thus providing interpretation for the content words. The proposed approach is demonstrated by extending the Attempto Controlled English (ACE) with polysemous and procedural constructs resulting in a more natural CNL named PAO covering narrative multi-domain texts.

## 1 Introduction

There are several sophisticated controlled natural languages (CNL), which cover relatively large subsets of English grammar, providing seemingly informal, high-level means for knowledge representation. CNLs typically support deterministic, bidirectional mapping to some formal language like first-order logic (FOL) or its decidable description logic (DL) subset [1], allowing integration with existing tools for reasoning, consistency checking or satisfiability model building.

Two widely accepted restrictions are used in CNLs to enable unambiguous construction of discourse representation structures: a set of interpretation rules for ambiguous syntactic constructions, and a monosemous lexicon — content words are not interpreted, they are treated as predicate identifiers whose meaning is defined only by FOL formulas derived from the text being analyzed[1]. While the first restriction limits only syntactic sophistication of a language, the second one causes essential communication limitations, as the natural language lexicon is inherently polysemous.

The problem of CNLs like ACE [3] is that although they include a rich lexicon of content words, this lexicon is purely syntactic and has no meaning or interpretation within the CNL itself. Currently it is left up to the CNL user to define (in the CNL text) the difference between the content words like "take" and "give" — this vital background knowledge is not part of the CNL. But the problem is even deeper — even users cannot know in advance all the possible meanings that might be associated with a particular word like "take": depending on the context "take" could mean creating a photo with a camera, depriving a person from something, helping someone to get home, or something else. This is where the need for polysemy support in CNL arises — only through polysemy it is possible to incorporate multi-domain background knowledge into the CNL.

[1] This is true for ACE and alike CNLs. There are other, non-deterministic CNLs such as CPL [2], which perform shallow semantic analysis to interpret content words.

The paper is organized in two parts — the first part (Sections 2, 3, 4) introduces the vital concepts of micro-ontology and multi-word unit to systematically cope (in a controlled manner) with noun polysemy in the classic CNLs such as ACE (formally, OWL DL[2] subset of ACE). In the second part of the paper (Sections 5, 6, 7) we go further and define a new kind of expressive CNL (named PAO) combining the declarative (static) and procedural (dynamic) background knowledge expressed in OWL DL ontologies and SPARQL/UL procedural templates respectively — this will allow to provide adequate support also for verbs and their polysemy. The new PAO CNL is illustrated on a simple fairytale fragment with the basic temporal and spatial expressions.

By this paper we also want to raise the general awareness about the role of polysemy as a gateway for incorporating the rich background knowledge into CNLs. The root cause of polysemy is that there is a "finite" set of words in the language, but there is an "unlimited" set of concepts in various domains or contexts [4] that might need to be named in order to be referenced. Although new words are invented over the time as well, this happens comparatively rarely and slowly — the new words have to be accepted and learned by the community. Therefore reuse of existing well-known words in different contexts is a common "workaround".

There are two main ways how words are reused [5]: metaphorically (relying on similarity, e.g. "mouse" for "computer pointing device") and metonymically (relying on relationship, e.g. "library" for "building of library"). There are also homonyms — words that "accidentally" have the same spelling, but their meanings and origins are unrelated — they can be seen more like exceptions.

Thus metaphoric and metonymic reuse of existing lexemes is unavoidable in the natural language, what can be summed up in a saying: language is a graveyard of "dead" metaphors [6]. Fortunately, it is observed [7] that the various senses of the same lexeme typically fall into different domains — therefore explicit identification of these domains is what enables the controlled polysemy described in this paper.

## 2 OWL DL Compliant Micro-ontologies as a Sense Inventory

A monosemous lexicon (terminology) could be appropriate for descriptions that verbalize single-domain knowledge, i.e., consistent OWL DL ontologies. However, even seemingly consistent descriptions might run into lexical ambiguities. A possible solution in such cases is to introduce ad-hoc lexemes by explicitly pointing out the specific meanings (e.g. "library-building" versus "programming-library"), but dependency on such ad-hoc naming makes the language unsystematic and, thus, user-unfriendly [9]. Instead, we will demonstrate how multi-word units (MWU) can be created systematically and largely automatically while dynamically merging together terminology of different domains.

---

[2] In this paper we use term OWL DL, a subset of OWL, OWL 1.1 and OWL 2.0, to avoid reliance on the specific OWL version [8].

Internally monosemous and consistent domain ontologies that follow lexically motivated naming convention we will call *micro-ontologies*[3]. By introducing the concept of micro-ontology we provide a systematic solution to the lexical ambiguity problem — instead of trying to make all content words globally unique through ad-hoc lexemes, we suggest splitting the background knowledge into multiple, relatively small and lexically unambiguous micro-domains[4] (see Fig.1). The benefit of this approach is its scalability to cover polysemous multi-domain terminology through a semi-automatic word sense partitioning procedure described in the next section.

<table>
<tr><td rowspan="5">TBox</td><td colspan="2">Micro-ontologies</td></tr>
<tr><td>Domain</td><td>Terminology (ontological text)</td></tr>
<tr><td>General</td><td>Every building is a physical_entity.<br>Every collection is an abstract_entity.<br>No physical_entity is an abstract_entity.</td></tr>
<tr><td>Building</td><td>Everything that has a roof is a building.<br>Every library is a building.<br>Every green_roof is a roof.</td></tr>
<tr><td>Programming</td><td>Everything that contains something is a collection.<br>Every library is a collection.<br>Every function is something.</td></tr>
<tr><td rowspan="3">ABox</td><td colspan="2">Assertions (factual texts)</td></tr>
<tr><td colspan="2">There is a library$_{\text{[building]}}$ that has a green_roof.</td></tr>
<tr><td colspan="2">AbsoluteValue is a function. [..] The library$_{\text{[programming]}}$ contains AbsoluteValue.</td></tr>
</table>

**Fig.1**. The three micro-ontologies verbalized in ACE illustrate emergence of polysemy for the lexeme “library” appearing in the terminological and assertional statements. The appropriate sense (namespace) has to be explicitly assigned to each utterance of “library” to create a consistent merged ontology.

By substituting a large, consistent ontology with numerous domain micro-ontologies we are inevitably introducing an ontology merging problem[5]. The problems of ontology merging (alignment, matching) and word sense disambiguation (WSD) are tightly intertwined and, in our view, the lack of definitive success in any of them is largely due to addressing these issues separately. Only by bringing both problems together it becomes possible to devise a method for semi-automatic *word sense partitioning* during micro-ontology merging process. The systematically partitioned word senses further enable semi-automatic *word sense disambiguation*.

---

[3] The proposed concept of micro-ontology bears some similarity to the concepts of environment or viewpoint in [10] and to the Cyc micro-theories [11], where all world-knowledge is split into narrow domain micro-theories. Meanwhile our word sense partitioning approach utilises standard OWL DL reasoning compatible with existing CNLs like ACE.

[4] We intentionally avoid discussion about the optimal size of micro-ontologies, as this is formally irrelevant. The choice might be between larger domains like Education, Sports, Finance or much smaller domains like FrameNet [12] frames.

[5] Although there is a vast literature on these issues (e.g. in [13]), majority of methods rely on shallow semantic similarity based on an external lexical taxonomy like WordNet [14]. Instead, we encourage to use the same OWL DL micro-ontologies as a sense inventory.

In the OWL DL subset of ACE-like CNLs, two kinds of statements can be differentiated [15]: terminological (ontological) and assertional (factual) ones, corresponding to the description logic TBox and ABox respectively. Ontological statements are those that introduce categories and describe relationships between them ("Every mouse is an animal"). Meanwhile factual statements are those talking about individuals belonging to specific categories ("The mouse$_{[computer]}$ X is connected_to the workstation Y").

In current CNLs both kinds of sentences usually are mixed into the same text: while populating facts one has to explicitly introduce also the ontology[6]. Instead, in our approach the two kinds of sentences must be strictly separated into *ontological texts* and *factual texts*. The rationale for separating these two kinds of texts is that ontological texts in our approach by definition are lexically monosemous within the scope of a single micro-ontology. Thus the WSD problem becomes limited only to the factual texts.

One has to remember that once a sufficient amount of background knowledge micro-ontologies are accumulated, the majority of the actual content will be lexically ambiguous multi-domain factual texts. By separating ontological and factual texts, our intention is to relieve an average CNL end-user from providing ontological statements (e.g. "Every library is a collection"), but rather allow the CNL end-user to concentrate on the factual content ("The library X contains a function Y"). The creation of domain micro-ontologies and their consistent merging (applying the semi-automatic word sense partitioning procedure described below) should be left to domain experts and knowledge engineers — as is the common practice already.

## 3 Word Sense Partitioning During Micro-ontology Merging

The formal semantics and decidability of OWL DL enables powerful means for what we call word sense partitioning. Namely, an OWL DL reasoner can automatically detect any formal inconsistencies (unsatisfiability) caused by incorrectly partitioned word senses during micro-ontology merging.

Note that in this section we consider only polysemy for class names (nouns) used within the micro-ontologies; property names are assumed to be globally monosemous among all micro-ontologies. This simplification will be justified in Section 5 where polysemy for verbs (predicates) will be introduced through procedural templates, making property names largely lexically invisible and, thus, less prone to problems of polysemy.

Assuming that all micro-ontologies reside in separate namespaces, a trivial approach to micro-ontology merging (possibly resulting in an inconsistent result) would be to add `owl:equivalentClass` axioms among all same-named classes from all micro-ontologies. This approach would have worked only if all lexemes used as class names were globally monosemous in all micro-ontologies.

In the presence of polysemous class names among different micro-ontologies, the actual merging procedure needs to avoid stating equivalence of same-named classes, if this would cause unsatisfiability of the merged ontology. Moreover, the problem is amplified by the possibility that the same name is used for classes, one of which actually is a subclass of the other (for example, “moon” in the astronomy micro-ontology might denote a satellite of any planet, while in the regular calendar micro-ontology it would more likely denote only the satellite of Earth). Such partitioning of same-named classes into separate meanings (senses) also requires introduction of unique, linguistically motivated sense identifiers to differentiate the distinct meanings of the same lexeme in the merged ontology; in the next section we will show how multi-word units can be used to address this issue systematically.

[6] Of course, ontological sentences can be manually reused with other factual texts pertaining to the same domain.

Assuming that in the given set of micro-ontologies there is a unique “conceptually correct” partitioning of senses for the polysemous class names, we would like to design a micro-ontology merging procedure that would find this partitioning automatically. Meanwhile one cannot have a free lunch: for the automatic procedure to work, the given set of micro-ontologies must satisfy the following strict conditions:

1. Micro-ontologies contain sufficient constraints to cause unsatifiability, if relation <X rdfs:subClassOf Y> is inserted between any pair of same-named classes, which are unrelated in the “conceptually correct” partitioning of senses.
2. Micro-ontologies contain sufficient constraints to cause unsatisfiability, if for any pair of subsuming same-named classes <X rdfs:subClassOf Y> in the “conceptually correct” partitioning of senses an opposite relation <Y rdfs:subClassOf X> is inserted.

Under the above conditions there is an automatic merging procedure, which correctly partitions meanings of the polysemous class names and creates the merged ontology (sense inventory). The merging procedure relies on the fact that the axiom <X owl:equivalentClass Y> is equivalent to the conjunction of axioms <X rdfs:subClassOf Y> and <Y rdfs:subClassOf X>. At the start of the merging procedure all same-named classes are partitioned, because they belong to distinct namespaces of respective micro-ontologies. To create a merged ontology the merging procedure attempts inserting rdfs:subClassOf relation (in both directions) between all pairs of same-named classes across all micro-ontologies and checks satisfiability of the resulting ontology with an OWL DL reasoner after every such insertion — only those rdfs:subClassOf insertions that are not causing unsatisfiability are preserved. The final stage of this procedure is to insert <X owl:equivalentClass Y> for all pairs of same-named classes for which both <X rdfs:subClassOf Y> and <Y rdfs:subClassOf X> relations have been inserted in the previous stage.

The described micro-ontology merging (sense partitioning) procedure will produce a correct result only if the above conditions (1) and (2) are met. There is no automatic way to tell if the produced merging result is conceptually correct or wrong — checking the merging result is the duty of the knowledge engineer, who will generally need to laboriously debug and fine-tune the micro-ontologies until the conditions (1) and (2) are met and the automatic merging procedure produces a conceptually correct sense partitioning. Nevertheless this semi-automated method is superior to manual creation of the merged ontology, because it relies on local validity of every included micro-ontology and thus scales with adding new micro-ontologies by potentially many domain experts — a goal hardly achievable otherwise.

The described micro-ontology merging procedure with word sense partitioning has to be performed only when the set of micro-ontologies is updated (this set of micro-ontologies can be considered a background knowledge part of CNL). This means that running of this procedure (and potential debugging of micro-ontologies) is generally the task of domain experts and knowledge engineers designing and combining the micro-ontologies into the sense inventory, and not the task of the CNL end-user, typically involved only with semi-automatic disambiguation in the factual texts against the given sense inventory, as illustrated in the next section.

## 4 Example of Controlled Polysemy in ACE-OWL

To illustrate how the described micro-ontology approach would fit into an existing CNL like OWL subset of ACE (we will call it ACE-OWL for short), let us consider the following example factual sentence, which could have been entered by the ACE-OWL end-user:

*A grandpa remembers* [*a*] *Germany that is not involved by* [*a*] *NATO.*

The use of indefinite articles together with proper names in the above example is somewhat artificial and is done only in order to guide the ACE parsing engine[7] to produce the below paraphrase where also proper names are treated as class names and not individuals. This is necessitated by the strict division of ontological background knowledge (introducing the terms, including proper names) and factual texts (using these terms for referring to specific individuals[8]). This approach is compensated by the uniform anaphora resolution detailed more in Section 7. Due to the strict separation of factual texts indefinite articles there could be assigned to proper names automatically and invisibly to the end-user.

*There is a grandpa X1.*
*The grandpa X1 remembers a Germany X2.*
*It is false that a NATO involves the Germany X2.*

To interpret the content of this factual text, a set of micro-ontologies describing the possibly polysemous meanings of appearing content words has to be defined in advance. Seven sample micro-ontologies are shown in Fig.2, describing some relevant background knowledge about geopolitics and people. The micro-ontologies themselves are presented in the regular ACE-OWL syntax [16]; the only "irregularity" is that the background knowledge is grouped into isolated, monosemous micro-ontologies, identified by the unique and explicitly stated namespaces. This allows to introduce polysemous meanings for content words such as "Germany" with the only restriction being that each meaning is described in a separate micro-ontology.

---

[7] See http://attempto.ifi.uzh.ch/site/tools/ (type all proper names in lower case with prefix `n:`).

[8] This is obvious for common nouns, but is often true also for proper nouns, because "Germany", for instance, could refer to "East Germany" or "West Germany".

| *Political map of Western Europe during the Cold War* | |
|---|---|
| (**we:**) http://m-ont.org/**ColdWarWesternEurope**.owl | |
| 1.1 | Every West_Germany is a country. |
| 1.2 | Every West_Germany is a Germany. Every Germany is a West_Germany. |
| 1.3 | Every NATO is an alliance. |
| 1.4 | Every West_Germany is involved by a NATO. |

| *Political map of Eastern Europe during the Cold War* | |
|---|---|
| (**ee:**) http://m-ont.org/**ColdWarEasternEurope**.owl | |
| 2.1 | Every Soviet_satellite_state is a country. |
| 2.2 | Every East_Germany is a Soviet_satellite_state. |
| 2.3 | Every East_Germany is a Germany. Every Germany is an East_Germany. |
| 2.4 | Every Warsaw_Pact is an alliance. |
| 2.5 | Every Soviet_satellite_state is involved by a Warsaw_Pact. |

| *Political map of Europe in 2007* | |
|---|---|
| (**eu:**) http://m-ont.org/**Europe2007**.owl | |
| 3.1 | Every federation is a country. |
| 3.2 | Every NATO_country is a country. |
| 3.3 | Every Germany is a federation. |
| 3.4 | Every Germany is a NATO_country. |

| *Bridging axioms for the political maps* | |
|---|---|
| (**b1:**) http://m-ont.org/b1.owl | |
| 4.1 | No {we,ee,eu,lg}:country that is involved by a we:NATO is involved by a ee:Warsaw_Pact. |
| 4.2 | Every eu:NATO_country is involved by a we:NATO. |
| 4.3 | No {we,ee,eu,lg}:country is an {we,ee,og}:alliance. |
| 4.4 | Every Prewar_Germany is a lg:Germany that is not involved by something that is a we:NATO or that is a ee:Warsaw_Pact. |

| *Map of official languages of European countries* | |
|---|---|
| (**lg:**) http://m-ont.org/**language**.owl | |
| 5.1 | Every German_speaking_country is a country. |
| 5.2 | Every Italian_speaking_country is a country. |
| 5.3 | Every Germany is a German_speaking_country. |
| 5.4 | Every Switzerland is a German_speaking_country. |
| 5.5 | Every Switzerland is an Italian_speaking_country. |

| *Taxonomy of organizational bodies* | |
|---|---|
| (**og:**) http://m-ont.org/**organization**.owl | |
| 6.1 | Every alliance is an organization. |
| 6.2 | Every alliance is a federation. Every federation is an alliance. |

| *Person name catalogue* | |
|---|---|
| (ps:) http://m-ont.org/**person**.owl | |
| 7.1 | Every man is a human. |
| 7.2 | Every woman is a human. |
| 7.3 | Every grandpa is a man. |

**Fig.2.** A set of sample micro-ontologies that describe small pieces of the changing political map of Europe. Few other micro-domains are sketched as well.

In cases where one micro-ontology needs to explicitly reference a class from another micro-ontology — a situation typical for "bridging" micro-ontologies (see `b1:` in Fig.2), a namespace identifier (or a list of identifiers) of the referenced remote ontology(-ies) must be included. We shall remind (see Section 3) that polysemy here is limited to class names only — all properties appearing in micro-ontologies are assumed to be monosemous and globally shared.

Besides designing micro-ontologies, the task of the knowledge engineer is also to ensure consistent merging of selected micro-ontologies by applying the algorithm described in Section 3. In case of a semantically incorrect matching of polysemous terms, the knowledge engineer must complement the micro-ontologies (or create a bridging ontology) with additional constraints.

A merger of the micro-ontologies given in Fig.2 is shown in Fig.3 where the (inconsistent) meanings of ambiguous words "Germany" and "federation" are split by the word sense partitioning algorithm. Although more restrictions could be added to separate today's "Germany" and "West_Germany" into different meanings, the current merger is sufficient for our demonstration.

Fig.3 also illustrates the use of multi-word units (MWU) to identify senses of the polysemous lexemes. MWU is a common technique used in natural language to differentiate meanings of polysemous lexemes (e.g. "computer mouse" to specify a meaning of the polysemous lexeme "mouse"). We use dashes to explicitly identify the two parts forming MWUs (e.g. "computer-mouse"). In our case the second part of a MWU is the original lexeme (including compound lexemes). As the first part of the MWU we recommend to use the name (namespace prefix) of one of the micro-ontologies, where this particular meaning appears. It should be noted that a MWU have to be created only in case of a polysemous lexeme.

The benefit of this approach is that self-explanatory MWUs can be created automatically, provided that micro-ontology names are linguistically motivated. Although it is questionable whether self-explanatory names can be provided for all micro-ontologies, this should be achievable in relatively isolated worlds. An alternative option is that the knowledge engineer manually defines an equivalent class with a more specialised name (term) which may be used for identification of the particular word sense.

Assuming that these tasks of the knowledge engineer are completed and the merged ontology with partitioned meanings of the polysemous lexemes is obtained, we can proceed with the parsing of the ACE-OWL factual text shown in the beginning of this section. The ACE parsing engine is used for the initial parsing of the input text, followed by a separate WSD step, during which the polysemous lexemes (in our example — "Germany") must be matched against the same-named classes (ignoring prefixes) of the merged ontology (Fig.3). Semi-automatic detection of a valid sense match (or possibly several valid sense matches) again is enabled by an OWL reasoner — a sense match is considered valid, if the resulting ontology (merged micro-ontologies and OWL statements corresponding to the factual input text) is still satisfiable. In case of the provided input text, only the following reading thus is found valid:

> *A grandpa remembers* [*a*] *ColdWarEasternEurope–Germany that is not involved by* [*a*] *NATO.*

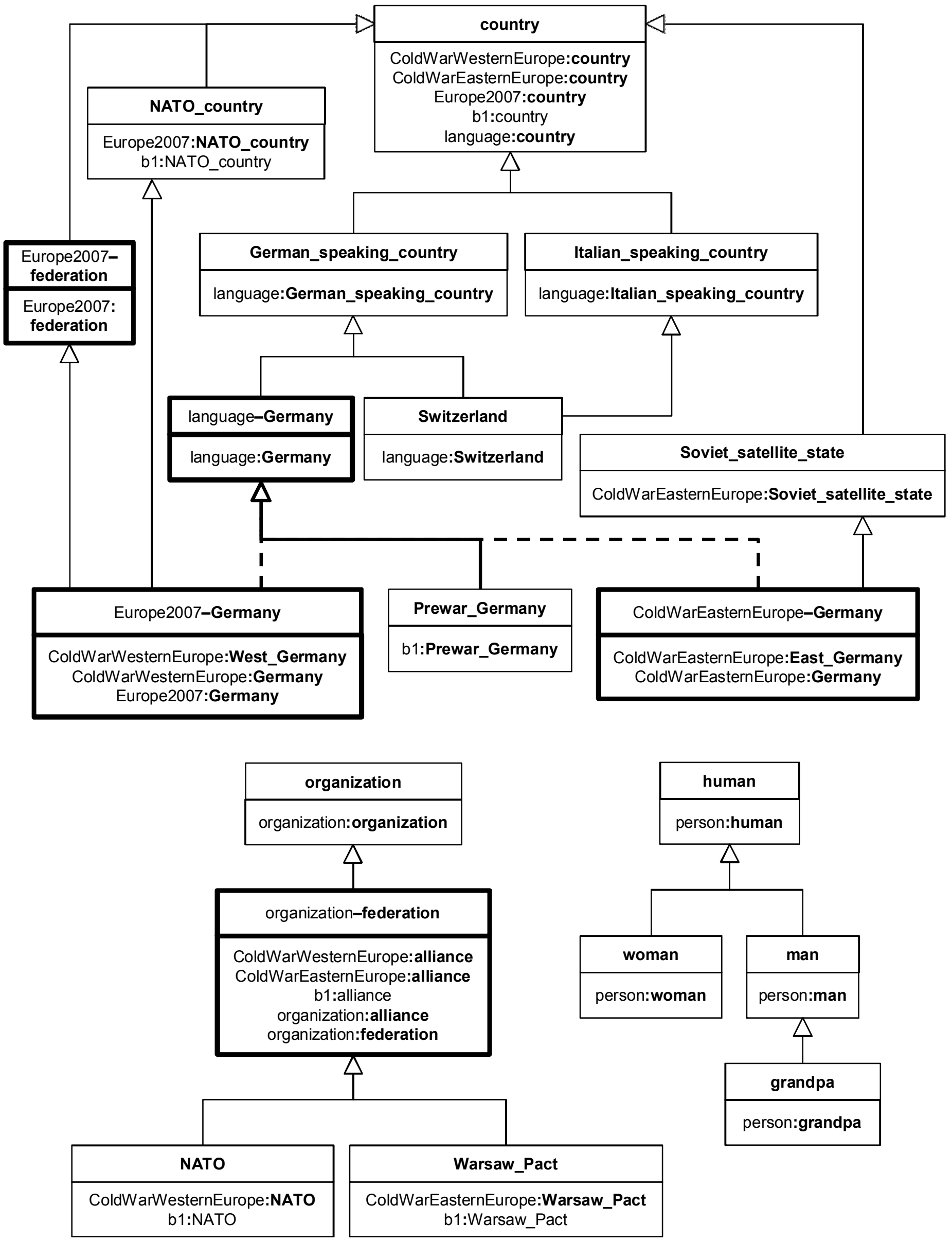

**Fig.3.** Merged micro-ontologies from Fig.2 with word senses partitioned as per algorithm described in Section 3. Each vertex in the graph represents a merged class: the lower section lists qualified names of the matched classes; the upper section shows name to be used for the merged class, which in case of polysemous lexeme may be a MWU. The dashed edges illustrate automatically detected subsumption relationships between senses of the same lexeme.

In general, the described WSD procedure might work only partially (leaving multiple options) due to insufficiently rich context provided in the factual input text.

Therefore there should be an alternative for the CNL end-user to choose the appropriate paraphrase manually — the automatic "guessing" technique would more likely be used only as a hinting engine for the CNL end-user, helping to select the meaning of a polysemous lexeme.

In contrast to the legacy macro-ontologies, micro-ontologies offer several significant advantages: (i) they do not impose a single consistent scheme, allowing many distinct points of view to co-exist; (ii) they can be seen as snapshots of some aspects of "reality", supporting non-stable and temporal entities; and (iii) they scale well — reality do not have to be compressed into a restricted number of lexical categories thus avoiding "signal losses" during conceptualisation.

## 5 PAO — a CNL with Polysemous and Procedural Constructs

In the previous part of the paper we tried to stay conservative and apply only minimum changes to the traditional ACE-OWL while adding polysemy for class-names and background knowledge support. In the following part of the paper we will try to stretch the limits and go beyond the traditional ACE capabilities in order to provide an adequate polysemy support also for verbs. We will do so by defining a new CNL named PAO (for Procedural ACE-OWL).

Besides polysemy support, the key difference of PAO is that it adds support also for procedural background knowledge in addition to the traditional declarative OWL ontology background knowledge. The distinction between procedural actions and declarative properties is neglected in OWL and related CNLs. Although temporal OWL ontologies are sometimes used, one has to remember that OWL as a subset of FOL has no native time concept and can introduce time only as part of the model; one has to go outside FOL and use procedural means to define model-changing-actions as will be illustrated with the execution trace in Section 7. In order to stay within the familiar OWL and RDF realm, SPARQL/UL[9] is an obvious choice for the procedural background knowledge component. In PAO we stay short of the full-fledged procedural language with IF-THEN-ELSE, GOTO and similar control features — these could be added, but are not necessary for the simple narrative text describing a plain sequence of actions such as a simple fairytale fragment used below to introduce the core PAO principles.

An additional rationale for combining OWL and SPARQL/UL is that only together they provide the expressive power equivalent to the regular relational database programming — ER-diagrams of relational database schemas (OWL equivalent) would be of little use without SQL language (SPARQL/UL equivalent). As we will see, this expressive power is adequate also for supporting narrative texts in a CNL.

PAO defines a natural and unambiguous translation from the controlled language input text into combination of OWL and SPARQL/UL. Formal definition of PAO is beyond the scope of this paper, but we will describe PAO by a detailed example covering its core features. Despite a rather different approach taken to define PAO, classic ACE-OWL technically will remain a subset of PAO.

---

[9] We will use the abbreviation SPARQL/UL for a union of SPARQL and SPARQL/Update — RDF query and update languages, now both part of SPARQL 1.1 [17].

# 6 Defining the Background Knowledge in PAO

The background knowledge necessary for interpretation of a CNL text can be described in the controlled language itself, or it can be imported from some other formalism. In PAO we permit both, but encourage the second approach due to potentially large size and complex structure of the background knowledge and, since also in real-life, background knowledge is often learned non-verbally through diagrams, examples or other means.

In PAO background knowledge consists of two parts — declarative OWL micro-ontologies (Fig.4) and procedural templates (Fig.6). The purpose of micro-ontologies (similarly to example in Section 4) is to define the *concept hierarchies* (OWL classes), their *relationships* (OWL properties) and *restriction axioms* (cardinality restrictions and others). The main design principle of micro-ontologies is to keep them lexically oriented to enable direct mapping of class (and optionally also property names) to the content words in the CNL text. Note that background knowledge is a pure TBox and never includes any OWL individuals (creation of ABox individuals will happen during anaphora resolution process) — for this reason in People micro-ontology there is introduced a subclass LittleRedRidingHood for all people (possibly just one) having this name — this approach enables uniform anaphora resolution mechanism to be used later. In PAO it is also required that each micro-ontology has a unique name (namespace). All class-names used in the micro-ontologies in Fig.4 have unique meaning (including the two meanings of lexeme “basket”) and, therefore, the automatic micro-ontology merging procedure from Section 3 here results in a simple union of the shown background knowledge input ontologies.

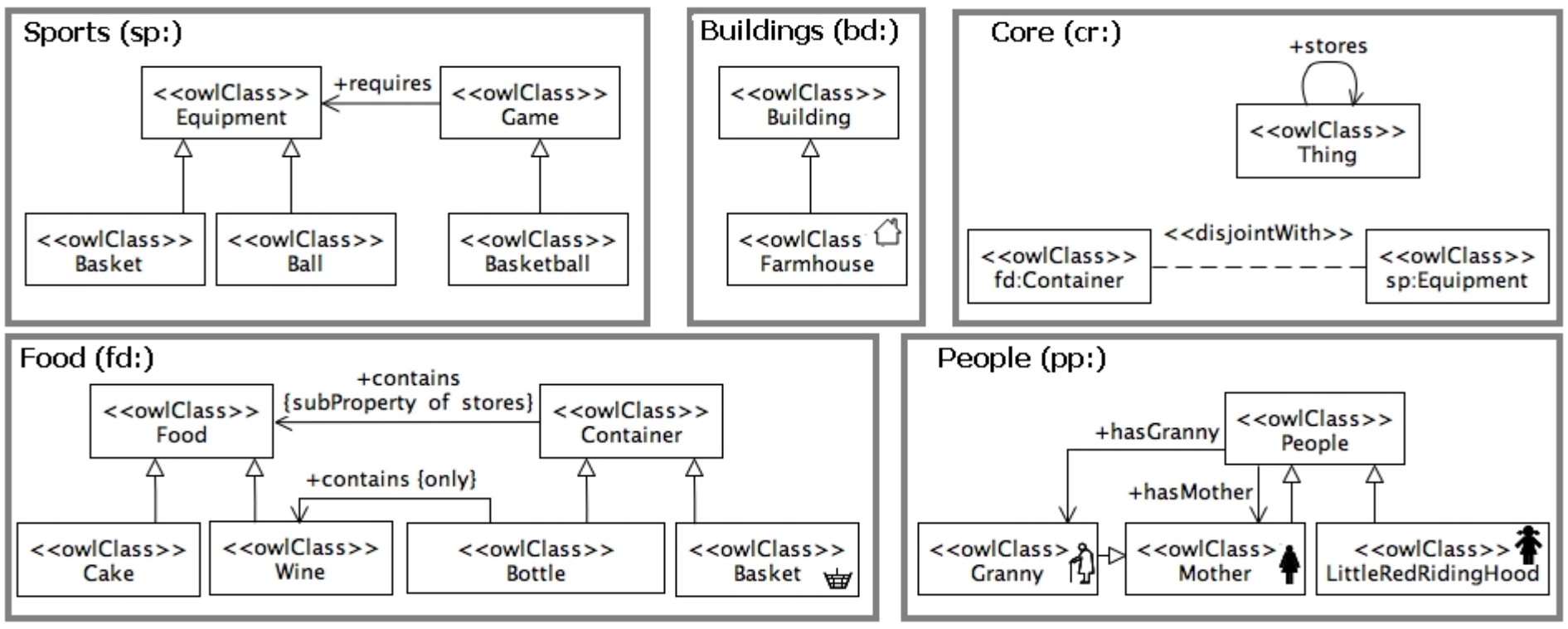

**Fig.4.** Background knowledge micro-ontologies in UML profile for OWL syntax.

In Fig.4 micro-ontologies are visualized according to the UML profile for OWL standard [18]. Alternatively, micro-ontologies may be defined verbally through ACE-OWL as illustrated in Fig.5 — this allows ACE-OWL ontological sentences (sentences dealing only with TBox) to remain a subset of PAO. The part of ACE-OWL dealing with ABox will be discussed in the next section.

*Every Basket is a Container.*
*Every Bottle is a Container.*
*Every Cake is a Food.*
*Every Wine is a Food.*
*Everything that contains something is a Container.*
*Everything that is contained by something is a Food.*
*Everything that is contained by a Bottle is a Wine.*
*If X contains Y then X stores Y.*

**Fig.5.** Example of ACE-OWL verbalization of Food micro-ontology from Fig.4.

The purpose of procedural templates background knowledge in Fig.6 is to provide a link between the action words (lexical units representing verbs) and their “meaning” in SPARQL/UL. As mentioned, the distinction between actions and properties is often neglected, but in PAO they are strictly separated: in PAO action is a non-ontological SPARQL/UL procedure, which *creates*/*deletes* individuals or *connects*/*disconnects* them through the predefined properties. PAO action, unlike binary OWL properties, has no arity restriction — it can link any number of arguments as is typical for verbs in natural language.

```
Procedure: Residence
  :parameters (?resident ?co-resident ?location)
  :precondition ()
  :effect (and(stores ?location ?resident)
          (stores ?location ?co_resident))
  :lexicalUnits (camp, inhabit, live, lodge, reside, stay)

Procedure: Removing
  :parameters (?agent ?source ?theme)
  :precondition (stores ?source ?theme)
  :effect (and(stores ?agent ?theme)
          (not(stores ?source ?theme)))
  :lexicalUnits (confiscate, remove, snatch, take, withdraw)

Procedure: Bringing
  :parameters (?agent ?goal ?theme)
  :precondition (and(stores ?agent ?theme)
                (stores ?a ?agent) (not(= ?a ?goal)))
  :effect (and(stores ?goal ?theme)(stores ?goal ?agent)
          (not(stores ?agent ?theme))
          (not(stores ?a ?agent)))
  :lexicalUnits (bring, carry, convey, drive, haul, take)
```

**Fig.6.** Procedural templates of background knowledge.

Syntactically a procedural template in PAO is a combination of elements inspired by FrameNet [12], Planning Domain Description Language (PDDL) [19] for situation calculus, and SPARQL/UL. The procedural template itself corresponds to FrameNet frame, the parameters section corresponds to FrameNet frame elements (only the actually used elements are included), and the lexical units section is a direct copy from FrameNet. Inclusion of precondition and effect sections in the procedural template is inspired by PDDL and has two-fold purpose: this is a compact representation of SELECT, INSERT, DELETE, MODIFY and WHERE patterns of the corresponding SPARQL/UL statement and at the same time it preserves compatibility with PDDL for planning purposes. Elements of planning will become necessary in the final steps of PAO interpretation described later. For word sense disambiguation purposes each procedural template must have a unique name.

Theoretically the precondition and effect sections of procedural templates could reference any class or property within any micro-ontology, but to achieve their maximum reusability, in PAO procedural templates are recommended to directly reference only universal properties with unrestricted domain and range — in Fig.4 they are depicted in a separate micro-ontology named “Core”, which contains also the necessary bridging axioms between the micro-ontologies. If a non-universal property needs to be manipulated by a procedural template, it can be defined as subproperty of some universal property (like “contains” is defined as a subproperty of the universal property “stores” in Fig.4).

The micro-ontologies and procedural templates shown in Fig.4 and Fig.6 are specifically crafted for the PAO example in the next section; for more realistic applications it would be necessary to create a much larger collections of micro-ontologies and procedural templates covering the whole lexicon and domain-knowledge of interest.

## 7 Example of PAO Text Processing

In this Section we are considering only narrative factual texts after the background knowledge (possibly including some ACE-OWL ontological text) has already been added into the system. Narrative texts in PAO have to be written in simple present tense to avoid complex event sequencing — the described events are assumed to be atomic and to occur sequentially as they are mentioned in the text. The following input text will be used in this paper to illustrate all processing stages of PAO.

> *LittleRedRidingHood lives in a farmhouse with her mother. She takes a basket from the farmhouse and carries it to her granny.*

This input text is ambiguous in at least three ways assuming that the background knowledge is limited to that defined in Fig.4 and Fig.6 in the previous section:

1. Anaphora “she” could refer to Little Red Riding Hood or to her mother,
2. “basket” could refer to the *food-basket* or *sports-basket*,
3. “take” could refer to *removing* or *bringing* procedure template.

In the first analysis stage PAO assists the user with rephrasing such ambiguous text into unambiguous paraphrase shown in Fig.7. PAO paraphraser automatically identifies ambiguities, generates the possible multi-word units for them (by prepending a micro-ontology or a procedural template name from the available background knowledge) and asks the user to select the correct alternative, which is then recorded into the paraphrase. Similarly PAO paraphraser asks the user to select the correct antecedent for unclear anaphors.

A. *Obj4 is a LittleRedRidingHood.*
B. *Obj4* ***lives*** *in obj8 with obj11.*
C. *Obj8 is a farmhouse.*
D. *Obj4 hasMother obj11.*
E. *Obj4* ***removing-takes*** *obj15 from obj8.*
F. *Obj15 is a* ***food-basket****.*
G. *Obj4 carries obj15 to obj25.*
H. *Obj4 hasGranny obj25.*

**Fig.7.** A disambiguated paraphrase of the input text.

Technically the PAO paraphraser is a syntactic parser complemented with matching rules towards a fixed list of atomic paraphrase patterns — this step is largely similar to ACE-OWL paraphrasing illustrated in Section 4 and to techniques used by automated FrameNet annotators [20, 21]. The selected multi-word units and anaphora antecedents can be inserted also in the original text to make it unambiguous, e.g. "She-LittleRedRidingHood removing-takes a food-basket from the…". Multi-word units are inserted only for the polysemous words.

Note that in the generated paraphrase in Fig.7 the statements A, C, D, F, and H are actually regular ACE-OWL factual statements about individuals within the background knowledge micro-ontologies. As such, these statements can be translated into corresponding OWL statements (actually RDF triples) by regular ACE-OWL means and these RDF triples then can be directly added into RDF database with a simple SPARQL/UL INSERT statement as shown in Fig.8. This procedure also ensures that PAO includes ACE-OWL factual statements (ABox) as a subset.

Meanwhile the procedural statements B, E, and G do not belong to ACE-OWL and require a procedural template from the background knowledge for their translation. The translation includes mapping of syntactic roles into procedural template parameters — such mapping techniques have been developed for automatic FrameNet annotation [20, 21] and their deterministic variation could be adapted also here. Next, the precondition and effect notation in the procedural template is translated into equivalent SPARQL/UL statement. In this second analysis stage paraphrase of the original input text gets converted into the sequence of SPARQL/UL statements shown in the first column of Fig.8.

Fig.8 contains also a second column which is filled in the third analysis stage. It contains SPARQL/UL statements, which implicitly follow from contents of the first column and the background knowledge in Fig.4 and Fig.6 — these statements are generated by the system automatically through OWL entailment and through PDDL-like planning over the preconditions and effects of the invoked procedural templates. The planning stage is needed here because in the input text some obvious intermediate steps of action might often be omitted and they need to be filled-in by planning to satisfy the procedural template preconditions — in our example for Little Red Riding Hood to be able to take a basket from the farmhouse, the basket had to be at the farmhouse in the first place. Here we leave entailment and planning explanation at the example level, but a more formal PAO definition would have to strictly limit the permitted extent of planning and entailment (for example, whether to allow here full OWL 2.0 reasoning over background knowledge micro-ontologies or to suffice with simpler RDFS entailment).

| | **EXPLICIT STATEMENTS** | **IMPLICIT STATEMENTS BY ENTAILMENT AND PLANNING** |
|---|---|---|
| A | INSERT {<obj4> <rdf:type><br><pp:LittleRedRidingHood>} | |
| **B** | **INSERT {<obj8> <stores> <obj4>.<br><obj8> <stores> <obj11>}** | |
| C | INSERT {<obj8> <rdf:type><br><bd:Farmhouse>} | INSERT {<obj8><br><stores> <obj15>}<br>*Inserted by planning because of procedural template precondition at step E.* |
| D | INSERT {<obj4> <pp:hasMother><br><obj11>} | INSERT {<obj11><br><rdf:type> <pp:Mother>}<br>*Entailed by range of the property pp:hasMother.* |
| **E** | **DELETE {<obj8> <stores> obj15}<br>INSERT {<obj4> <stores> <obj15>}** | |
| F | INSERT {<obj15> <rdf:type><br><fd:Basket>} | |
| **G** | **DELETE {<obj4> <stores> <obj15>.<br>?a <stores> <obj4>}<br>INSERT {<obj25> <stores> <obj15>.<br><obj25> <stores> <obj4>}<br>WHERE {?a <stores> <obj4>.<br>FILTER (?a != <obj25>)}** | |
| H | INSERT {<obj4> <pp:hasGranny><br><obj25>} | INSERT {<obj25><br><rdf:type> <pp:Granny>}<br>*Entailed by range of the property pp:hasGranny.* |

**Fig.8.** Generated SPARQL/UL statements (procedural statements are in bold).

The fourth analysis stage is to execute the SPARQL/UL statement sequence shown in Fig.8 and to generate the RDF database content trace of such execution — Fig.9 shows the resulting stepwise RDF database content trace. Such content trace could technically be stored as a sequence of RDF named graphs, along with an additional named graph storing the background knowledge micro-ontologies.

The generated RDF database content trace is the final result of PAO text analysis — this trace is the actual discourse conveyed by the PAO input text. In the right column of Fig.9 the discourse is optionally visualized also as a sequence of graphic scenes — similarly to text-to-scene animation approach described in [22]. These visualizations can be generated automatically from the graphic icons provided for OWL classes in the background knowledge (Fig.4 includes the necessary icons); relations are visualized as labelled arrows or alternatively spatial relations like

| | | |
|---|---|---|
| *A* | *<obj4> <type> <LittleRedRidingHood>.* | 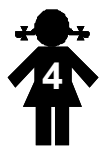 *Obj4 is a LittleRedRidingHood.* |
| *B* | *<obj4> <type> <LittleRedRidingHood>.*<br>*<obj8> <stores> <obj4>.*<br>*<obj8> <stores> <obj11>.* | 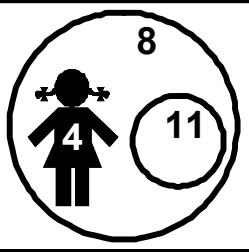 *Obj4 lives in obj8 with obj11.* |
| *C* | *<obj4> <type> <LittleRedRidingHood>.*<br>*<obj8> <stores> <obj4>.*<br>*<obj8> <stores> <obj11>.*<br>*<obj8> <type> <farmhouse.*<br>*<obj8> <stores> <obj15>* | 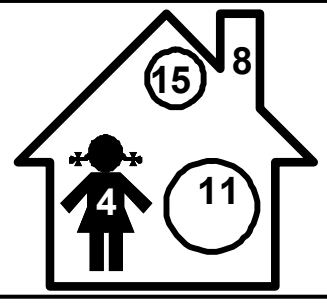 *Obj8 is a farmhouse.* |
| *D* | *<obj4> <type> <LittleRedRidingHood>.*<br>*<obj8> <stores> <obj4>.*<br>*<obj8> <stores> <obj11>.*<br>*<obj8> <type> <farmhouse>.*<br>*<obj4> <hasMother> <obj11>.*<br>*<obj11> <type> <mother>.*<br>*<obj8> <stores> <obj15>* | 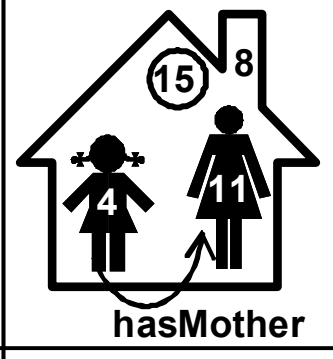 *Obj4 hasMother Obj11.* |
| *E* | *<obj4> <type> <LittleRedRidingHood>.*<br>*<obj8> <stores> <obj4>.*<br>*<obj8> <stores> <obj11>.*<br>*<obj8> <type> <farmhouse>.*<br>*<obj4> <hasMother> <obj11>.*<br>*<obj11> <type> <mother>.*<br>*<obj4> <stores> <obj15>* | 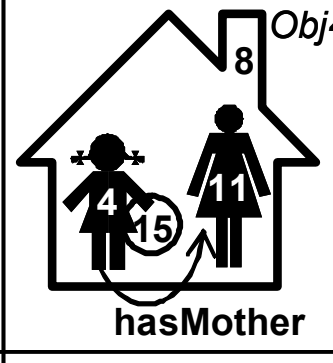 *Obj4 removing-takes obj15 from obj8.* |
| *F* | *<obj4> <type> <LittleRedRidingHood>.*<br>*<obj8> <stores> <obj4>.*<br>*<obj8> <stores> <obj11>.*<br>*<obj8> <type> <farmhouse>.*<br>*<obj4> <hasMother> <obj11>.*<br>*<obj11> <type> <mother>.*<br>*<obj4> <stores> <obj15>*<br>*<obj15> <type> <food-basket>* | 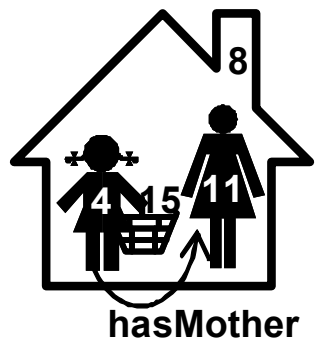 *Obj15 is a food-basket.* |
| *G* | *<obj4> <type> <LittleRedRidingHood>.*<br>*<obj25> <stores> <obj4>.*<br>*<obj8> <stores> <obj11>.*<br>*<obj8> <type> <farmhouse>.*<br>*<obj4> <hasMother> <obj11>.*<br>*<obj11> <type> <mother>.*<br>*<obj25> <stores> <obj15>.*<br>*<obj15> <type> <food-basket>.* | 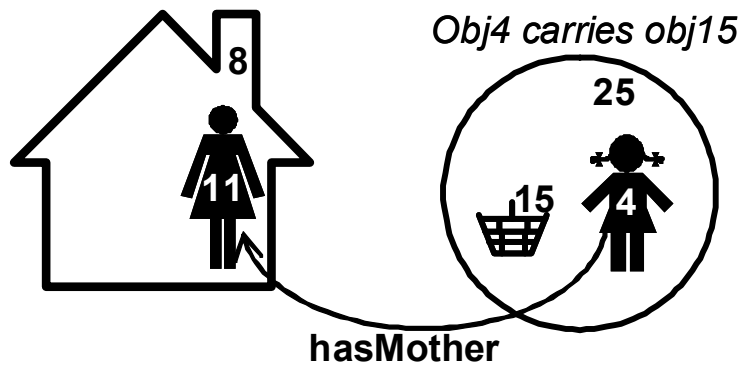 *Obj4 carries obj15 to obj25.* |
| *H* | *<obj4> <type> <LittleRedRidingHood>.*<br>*<obj25> <stores> <obj4>.*<br>*<obj8> <stores> <obj11>.*<br>*<obj8> <type> <farmhouse>.*<br>*<obj4> <hasMother> <obj11>.*<br>*<obj11> <type> <mother>.*<br>*<obj25> <stores> <obj15>.*<br>*<obj15> <type> <food-basket>.*<br>*<obj4> <hasGranny> <obj25>.*<br>*<obj25> <type> <granny>* | 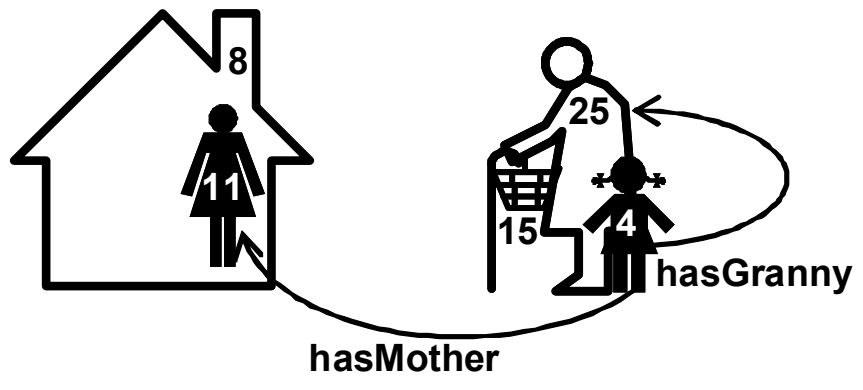 *Obj4 hasGranny Obj25.* |

**Fig.9.** RDF database content trace and its optional visualization.

“stores” can be visualized as graphic inclusion. These visual scenes highlight the similarity of PAO analysis result to the dynamic scene likely imagined by a human reader incrementally reading the same input text.

The constructed RDF database trace in Fig.9 can further be used to answer queries about the input text, for example:

1. *Who delivered a basket to a granny?*
2. *Did LittleRedRidingHood visit her granny?*
3. *Where initially was the basket?*
4. *When did the granny got the basket?*

To see how these queries could be answered through the constructed RDF database content trace, they first need to be disambiguated and with the same PAO techniques translated into the following SPARQL queries extended with the non-standard trace step identification in the WHERE-AT-STEP section (technically this non-standard trace step identification could be implemented through more lengthy RDF named graphs manipulation):

```
1. SELECT ?x
   WHERE-AT-STEP(?n) {?w <stores> ?x. ?x <stores> ?y.}
   WHERE-AT-STEP(?n+1) {
     ?z <stores> ?x. ?z <stores> ?y.
     ?y <rdf:type> <fd:Basket>.
     ?z <rdf:type> <pp:Granny>}
2. SELECT * WHERE-AT-STEP(any) {
     ?z <stores> ?x.
     ?x <rdf:type> <pp:LittleRedRidingHood>.
     ?z <rdf:type> <pp:Granny>}
3. SELECT ?x WHERE-AT-STEP(min) {
     ?x <stores> ?y.
     ?y <rdf:type> <fd:Basket>}
4. SELECT ?n WHERE-AT-STEP(?n) {
     ?y <stores> ?x.
     ?x <rdf:type> <fd:Basket>.
     ?y <rdf:type> <pp:Granny> }
```

The answers produced by these queries on the RDF trace in Fig.9 would be:

```
1. ?x = obj4
2. yes
3. ?x = obj8
4. ?n = H
```

These very technical SPARQL answers could afterwards be rendered into more verbose answers:

1. *LittleRedRidingHood* [*delivered a basket to granny*].
2. *Yes* [*, LittleRedRidingHood visited granny*].
3. [*Basket initially was*] *in the farmhouse.*
4. *In step H* [*, when LittleRedRidingHood brought the basket to granny*].

Although we have not described the question answering process here in detail, these examples provide an overview of PAO potential for factual and temporal question answering over narrative input texts.

## 8 Conclusion

We have shown that noun polysemy can be introduced into CNLs in a controlled manner through dividing the underlying terminological ontology into monosemous micro-ontologies. This not only provides a consistent naming for different senses of the same word through prefixing it with the source micro-ontology name, but also allows automatic OWL DL reasoning techniques to be employed for identifying and merging the non-conflicting uses of the same word in different micro-ontologies. The key benefit of this approach is that terminology of a particular domain is localised to the corresponding micro-ontology and the same words in this way can be freely reused in other micro-ontologies covering other parts of the background knowledge. In this way it becomes possible to bundle the rich and extensible background knowledge about content words within the CNLs like ACE.

Meanwhile to properly cover also verb (action, property) polysemy, we had to introduce a new CNL named PAO. The described PAO controlled language is only a rather simple attempt to exploit the rich declarative and procedural background knowledge in controlled natural language. We are quite pleased to have managed to include ACE-OWL as a proper subset of PAO thus achieving a complete complementary integration of procedural and declarative approaches. Also the briefly mentioned optional visualization of PAO discourse is a tempting area for further exploration — inversion of the mentioned visualization technique could lead to a vision simulation grounded in the same ontological and procedural background knowledge.

An obvious limitation of the presented PAO language is its treatment of time only as a linear sequence of events mentioned in the input text. A richer time conceptualization is generally needed, including hypothetical, parallel and negated events [23] to handle texts like "Mother told LittleRedRidingHood to go directly to the granny's house and not to engage in conversations with strangers".

Nevertheless, the described approach to polysemy in our view provides a new insight into how background knowledge can be systematically added into CNLs and how multiple kinds of background knowledge (procedural and declarative) extend CNL expressivity.

**Acknowledgements.** The underlying project is funded by the National Research Program in Information Technologies (Project No. 2). We also thank the reviewers of the CNL 2009 workshop proceedings for their valuable comments.